\documentclass{article}

\PassOptionsToPackage{numbers, compress}{natbib}
\usepackage[final]{nips_2016}

\usepackage[utf8]{inputenc} 
\usepackage[T1]{fontenc}    
\usepackage{url}            
\usepackage{booktabs}       
\usepackage{amsfonts}       
\usepackage{nicefrac}       
\usepackage{microtype}      
\usepackage{amssymb}
\usepackage{amsmath}
\usepackage{amsthm}
\usepackage{algorithm}
\usepackage{algpseudocode}
\usepackage{array}
\usepackage{enumitem}
\usepackage{graphicx}
\usepackage{lscape}
\usepackage{multirow}
\usepackage{parskip}
\usepackage{tabto}
\usepackage{mathtools}
\usepackage{caption}
\usepackage{subcaption}

\title{Imitation Learning with Recurrent Neural Networks}

\author{
  Khanh Nguyen\\
  Department of Computer Science\\
  University of Maryland\\
  College Park, MD 20742 \\
  \texttt{kxnguyen@cs.umd.edu} \\
}

\begin{document}

\maketitle

\begin{abstract}
  We present a novel view that unifies two frameworks that aim to solve sequential prediction problems: learning to search (L2S) and recurrent neural networks (RNN). We point out equivalences between elements of the two frameworks. By complementing what is missing from one framework comparing to the other, we introduce a more advanced imitation learning framework that, on one hand, augments L2S's notion of search space and, on the other hand, enhances RNNs' training procedure to be more robust to compounding errors arising from training on highly correlated examples.  
\end{abstract}

\section{Introduction}

Tackling sequential prediction problems is a formidable challenge in machine learning. 
The number of possible labels in a sequential prediction problem increases exponentially with the length of the sequence, making considering all label configurations intractable. 
On the other hand, a naive supervised learning approach that makes an independent prediction at each time step suffers seriously from compounding errors since the input observations correlate and thus violate the identically independent distributed assumption required by any supervised approach. 

Learning to search (L2S) \cite{collins2004incremental, daume2009search, daume2005learning, ross2011reduction, doppa2014hc, ross2014reinforcement, krishnamurthy2015learning} employs a similar approach but is immune to the shortcomings of the naive supervised approach. L2S algorithms reduce a sequential prediction problem to learning a policy to traverse in a search space with minimum cost. 
Under mild assumptions, L2S algorithms such as SEARN or DAgger \cite{daume2009search, ross2011reduction} generally guarantee that compounding errors grow linearly with trajectory lengths, which is theoretically as good as applying the supervised approach on identically independent distributed data.
 
Like other machine learning algorithms, L2S algorithms also suffer from the problem of data sparsity. 
At testing time, if the algorithms encounter a previously unseen state, they are likely to predict poorly. 
We argue that one solution to this problem is being able to capture the similarities between states, so that unseen states can still be approximated by other seen states. 
Previous works on L2S define the search space representation by specifying a hand-crafted feature extraction function, which is usually discrete and static, thus capture similiarities poorly. 
We introduce a \emph{deep learning} approach to learning a \textit{continuous representation} of the search space. In such a search space, an unseen state can be approximated by its seen neighbors. 
In our approach, the search space's representation is learned jointly with the optimal policy, so it is optimized to facilitate the prediction-making process. 

Specifically, we show how to formulate a standard L2S problem from elements of a recurrent neural network. We demonstrate that, learning the search space representation and search policy using an L2S algorithm such as DAgger is equivalent to training a recurrent neural network model by \textit{scheduled sampling} \cite{bengio2015scheduled}. Hence, our method can be considered as both an augmentation to L2S algorithms and a novel training algorithm for recurrent neural networks.

\section{Problem setting}
\subsection{Learning to Search}
\label{sec:l2s}

Learning to search (L2S) frames a sequential predicting process as having a learning agent traversing through states in a search space $S$. 
There are two special subsets of states in that space: the set of start states $S_{start}$ and the set of final states $S_{final}$. 
The learning agent commences at one of the states in $S_{start}$. At each time step, from the current state, it chooses an action from a set of actions, $A$, and transitions to a next state following a transition function $\Psi : S \times A \rightarrow S$. 
A \textit{complete trajectory} of length $T$ is a sequence of states and actions that terminates at a state in $S_{final}$:
$$ s_0, a_0, s_1, a_1, \dots, a_{T - 1}, s_{T} \in S_{final}$$ where $s_i = \Psi \left( s_{i - 1}, a_{i - 1} \right)$ for $i > 0$.

To determine which action should be taken next from a state, the learning agent execute a policy $\pi : S \rightarrow A$, which maps a state to an action. 
L2S is a form of imitation learning, thus the goal is to learn the ``best'' policy $\hat{\pi}$ that mimics a reference policy $\pi^{*}$.
The reference policy is provided by an expert at training time, but not necessarily optimal. 
To formalize the learning goal, we define a loss function $l_{\pi} : S \rightarrow R$, whose value at a state $s$ is the cost of taking a complete trajectory starting at $s$, then following $\pi$ instead of folowing $\pi^{*}$. 
The best policy miminizes the expected loss over its induced state distribution:
\begin{equation} \pi^* = \arg \min_{\pi} \mathbb{E}_{s \sim d_{\pi}} \left[ l_{\pi}(s) \right] \end{equation} where $d_{\pi}$ is the state distribution induced by policy $\pi$.

In practice, the raw representation of the search state may not be helpful for learning a good policy. 
For example, in natural language processing problems, each search state is an instance of raw text; those search states are usually too sparsely distributed to capture any meaningful relationships. 
In those cases, to reduce sparsity, we use a feature extraction function, $\Phi : S \rightarrow S_{\Phi}$, to project the raw representation of a search state onto a feature space, $S_{\Phi}$. 
For simplicity, we re-define the policy to directly take input from the feature space, i.e. $\pi : S_{\Phi} \rightarrow A$. 

\subsection{Recurrent neural networks}

Recurrent neural networks are a variant of neural networks which targets sequential prediction problems. They are widely used to tackle supervised problems involving learning a mapping from an input sequence $x = \{x_t \}$ to an output sequence $y = \{y_t\}$.

During a feed-forward pass, a recurrent neural network maintains hidden representations of the input observations it has seen. Specifically, at the beginning, an initial input observation, $x_0$, is fed into the network. 
During time step $t > 0$, the current input observation $x_{t}$ and the previous hidden vector $h_{t - 1}$ are composed with a non-linear function $f_{\theta}$, parametrized by $\theta$, to produce a new hidden vector $h_t$:
\begin{equation}
h_t = 
\begin{cases} 
  x_0 & \text{if } t = 0;
  \\
  f_{\theta} \left( h_{t - 1}, x_{t} \right) & \text{otherwise.} 
\end{cases}
\label{eqn:seqseq}
\end{equation}

A discrete distribution over $y_t$ can be obtained by applying a softmax function on $h_t$:
\begin{equation} P_{\theta} \left( y_t \mid h_t\right) = \text{softmax}(h_t) \end{equation}

from which we can infer a prediction $\hat{y}_t$ about $y_t$:
\begin{equation} 
  \hat{y}_t = \arg \max_{y} P \left( y_t = y \mid h_t ; \theta \right) 
  \label{eqn:inference}
\end{equation}

The network is typically trained to maximize the log-likelihood of the output sequences given the input sequences of a training set $\mathcal{D} = \left\{ \left( x^{(i)}, y^{(i)} \right) \right\}$:

$$ \theta^{*} = \arg\max_{\theta} \ \log \sum_{(x^{(i)}, y^{(i)}) \in \mathcal{D}} P_{\theta} \left( y^{(i)} \mid x^{(i)} \right)$$

where 
$$ \log P_{\theta} \left( y \mid x \right) = \sum_t \log P_{\theta} \left( y_t \mid h_t \right) $$

\section{Learning to search under recurrent neural network framework}

Surprisingly, the RNN framework naturally fits the L2S framework without requiring any substantial modification.
In this section, we show how to convert a supervised learning problem on an RNN into a standard L2S problem. Concretely, we specify the fundamental components of a L2S problem (search space, states, action, transition function, policy) using fundamental components of an RNN (input, output, hidden representations, non-linear function).

The main idea is to view the hidden representations of an RNN as states in a continuous search space and the non-linear transformation as a means of transitioning from states to states in that space. Formally, we denote the dimension of the hidden vectors of the RNN by $H$ and set the search space to be $R^H$. A forward pass through an RNN is considered as traversing over states $h_t$ in this space. At the beginning of time, the learning agent visits a start state, which varies depending on the input sequence. This state is computed by a feature extraction function $\Phi$. The output sequence $y$ is deemed as a series of actions the learning agent takes to transition to a next state from a current state. To construct a transition function, we modify the RNN's non-linear function $f_{\theta}$ to take into account the previous prediction.

\begin{algorithm}[t]
  \begin{algorithmic}[1]
    \Function{Traverse}{$x, y, \Phi, f_{\theta}$}
    \State Compute start state $h_0 = \Phi(x)$.
    \For {each $(x_t, y_t)$}
    \State Take action $y_t$.
    \State Compute next state $h_t = f_{\theta}(h_{t - 1}, y_{t - 1})$.
    \EndFor
    \EndFunction
  \end{algorithmic}
  \caption[2]{\label{alg:traverse} Training algorithm for a seq2seq model using mini-batch DAgger.}
\end{algorithm}

Algorithm \ref{alg:traverse} summarizes the traversing procedure. We drop the parameter $x_t$ of $f_{\theta}$ to be consistent with the definition of the transition function in the L2S framework. However, in practice, we can allow $f_{\theta}$ to take on $x_t$, i.e. $h_t = f_{\theta}(h_{t - 1}, y_{t - 1}, x_t)$. This form of function is widely used in sequence-to-sequence models \cite{sutskever2014sequence}. $x_t$ can be interpreted as new information coming from the environment when arriving at a new state. Hence, it plays an important role in the computation of the next states -- a fact not modeled explicitly in L2S framework but still practically implemented. The feature extraction function $\Phi$ is any function that encodes an input sequence or a part of it. For example, for sequence-to-sequence models, $\Phi$ is an encoding RNN.   

Given that the model hyperparameters are fixed, a policy is equivalent to a configuration of model parameters. Hence, we abuse the notation $\theta$ to denote both a parameter vector and a policy function. 

A learned policy is the inference function of the model:
\begin{equation}
  \hat{\theta}(h_t) = \arg \max_{y} P(y_t = y \mid h_t; \theta) 
\end{equation}

The reference policy is the policy that always outputs the true labels:
\begin{equation}
  \theta^{*}(h_t) = y_t 
\end{equation}

\section{Policy learning}

The regular training procedure of RNNs treats true labels $y_t$ as actions while making forward passes. 
Hence, the learning agent follows trajectories generated by the reference policy rather than the learned policy. 
In other words, it learns:
\begin{equation} \hat{\theta}^{sup} = \arg \min_{\theta} \mathbb{E}_{h \sim d_{\pi^{*}}} \left[ l_{\theta}(h) \right] \end{equation}

As mentioned in Section \ref{sec:l2s}, however, our true goal is to learn a policy that minimizes errors under its own induced state distribution:
\begin{equation} \hat{\theta} = \arg \min_{\theta} \mathbb{E}_{h \sim d_{\theta}} \left[ l_{\theta}(h) \right] \end{equation}

At testing time, we no longer have access to true labels but only predicted labels. In other words, we have to replace the reference policy by our learned policy.  
If the learned policy does not mimic the reference policy perfectly, which is often the case due to limited training examples, the discrepancy between the two policies causes the model to suffer severely from compounding errors. 
The learning agent may run into a state it has never seen before and take an wrong action that leads to a completely unrecoverable trajectory. 
This is exactly the problem we encountered with using a naive supervised approach on non-identically independently distributed data. Here the dependencies between training examples are even more explicit because each hidden state is directly computed from the hidden state from the previous time step. 

In order to effectively learn a policy that performs well at testing time, we apply L2S algorithms to training RNNs.
Algorithm \ref{alg:train} presents an online DAgger approach \cite{ross2011reduction} to training RNNs, which has been proposed previously under the name \emph{scheduled sampling} \cite{bengio2015scheduled}. This method is shown to improve performances of RNN sequence-to-sequence models on various tasks such as image captioning, syntactic parsing and speech recognition. The training procedure does not differ much from the regular mini-batch training procedure for RNNs. The only difference is that, at each time step, the model \emph{stochastically} determines whether to use the true label or the predicted label to compute the next hidden state. The probability of using true labels should decay over time so that, at the end of training, the model is trained almost entirely on its own policy's state distribution. It is important to note that predicted labels only serve to compute the search states. The examples used for updating model parameters still carry true labels. In other words, the learning agent traverses in the search space using its learned policy while learning what an expert would do at the visited states.      

\begin{algorithm}[t]
  \begin{algorithmic}[1]
    \Function{Train}{$N, \alpha$}
    \State Intialize $\alpha = 1$.
    \State Initialize model parameters $\theta$.
    \For {$i = 1 .. N$}
    \State Set $\alpha = \alpha \cdot p$.
    \State Randomize a batch of labeled examples.
    \For {each example $(x, y)$ in the batch}
    \State Initialize $h_0 = \Phi(X)$.
    \State Initialize $\mathcal{D} = \left\{ (h_0, y_0) \right\}$.
      \For {$t = 1 \dots |Y|$}
      \State Uniformly randomize a floating-number $\beta \in [0, 1)$. 
        \If {$\alpha < \beta$}
        \State Use true label $\tilde{y}_{t - 1} = y_{t - 1}$
        \Else
        \State Use predicted label: $\tilde{y}_{t - 1} = \arg \max_y P(y \mid h_{t - 1}; \theta)$.
        \EndIf
      \State Compute the next state: $h_t = f_{\theta}(h_{t - 1}, \tilde{y}_{t - 1})$.
      \State Add example: $\mathcal{D} = \mathcal{D} \cup \left\{ (h_t, y_t) \right\}$.
    \EndFor
    \EndFor
    \State Online update $\theta$ by $\mathcal{D}$ (mini-batch back-propagation). 
    \EndFor
    \EndFunction
  \end{algorithmic}
  \caption[2]{\label{alg:train} Training algorithm for a seq2seq model using mini-batch DAgger.}
\end{algorithm}

\section{Discussion}

The RNN formulation of L2S offers a perspicuous interpretation of the notion of ``search space'' in the L2S framework. 
As implied in Equation \ref{eqn:seqseq}, a search state encodes the action sequence taken to reach it. 
Therefore, rather being a unknown given universe as described in previous works on L2S, the search space in the RNN formulation explicitly represents a composite of encodings of all possible action sequences. 
Besides that, embedding the search space in a vector space also allows us to reason about it geometrically. 
For instance, we can explicitly measure the similarity between two states by computing the distance between their representation vectors with respect to a geometric metric such as cosine distance. 

More importantly, learning a continous search is an effective solution to the state sparsity problem and thus can potentially improve performance of L2S algorithms. 
A adaptive, dense continuous representation is superior to a static, sparse hand-crafted feature representation in two ways. 
First, it is capable of disentangling hidden non-linear dimensions of the sparse representation and providing a rigorous sense of similiarities between states. 
Capturing similarity is crucial to dealing with sparsity, as observed in many other machine learning problems where neural networks yield superior performances. 
In the context of imitation learning, suppose that the learning agent encounters an previously unknown state and does not know what to do, the ability to relate to what it would do in ``similar'' known states can still offer an approximated view of the current situation. 
Second, through back-propagation training, the search space's representation is optimized simultaneously with the learned policy.
Hence, beside having control over the learned policy, the learning agent also gains control over the distribution of states in the search space and can position those states in a way that is beneficial for its decision-making process. 
This allows the learning agent to improve its predicting capability not only by learning to analyze his situations better, but also by learning to realize his situations better. 

\bibliographystyle{plain}
\bibliography{nips_2016}

\end{document}